\documentclass[lettersize,journal]{IEEEtran}
\usepackage{amsmath,amsfonts}
\usepackage{algorithmic}
\usepackage{algorithm}
\usepackage{textcomp}
\usepackage[caption=false,font=normalsize,labelfont=sf,textfont=sf]{subfig}
\usepackage{textcomp}
\usepackage{stfloats}
\usepackage{url}
\usepackage{verbatim}
\usepackage{graphicx}
\usepackage{multirow}
\usepackage{microtype}
\usepackage{rotating}
\usepackage{tabularx}
\usepackage{booktabs} 
\usepackage{array}    
\usepackage{multirow} 
\usepackage{caption}
\usepackage{pdflscape}
\usepackage{array}
\usepackage{enumitem}
\usepackage{graphicx}
\usepackage{makecell}
\usepackage{booktabs} 
\usepackage{makecell} 
\usepackage{multirow}
\usepackage{newunicodechar}
\newunicodechar{：}{:}  
\usepackage{amssymb} 
\usepackage{pifont}
\usepackage[numbers]{natbib}
\newcommand{\cmark}{\ding{51}} 
\newcommand{\xmark}{\ding{55}} 
\renewcommand{\arraystretch}{1.5} 

\begin{document}

\title{Cooperative Visual-LiDAR Extrinsic Calibration Technology for Intersection Vehicle-Infrastructure： A review\\}
\author{Xinyu Zhang$^{1}$, Yijin Xiong$^{1,2}$, Qianxin Qu$^{1}$, Renjie Wang$^{1}$, Xin Gao$^{1,2}$, Jing Liu$^{2}$, Shichun Guo$^{1}$, Jun Li$^{1}$

 \thanks{$^{1}$Xinyu Zhang, Yijin Xiong, Qianxin Qu, Renjie Wang, Xin Gao, Shichun Guo and Jun Li are with the school of Vehicle and Mobility, Tsinghua University, Beijing, China.}	
		
 \thanks{$^{2}$Yijin Xiong, Xin Gao, Jing Liu are with the Computer Science and Technology, China University of Mining and Technology-Beijing, Beijing, China. 
 		(e-mail: bqt2000405025@student.cumtb.edu.cn; bqt2000405024@student.cumtb.edu.cn; zgy@cumtb.edu.cn; bqt2200405021@student.cumtb.edu.cn)
 }	
\thanks{This work was supported by the National High Technology Research and Development Program of China under Grant No. 2018YFE0204300, and the National Natural Science Foundation of China under Grant No. 62273198, U1964203.}
 }



\maketitle

\begin{abstract}
In the typical urban intersection scenario, both vehicles and infrastructures are equipped with visual and LiDAR sensors. By successfully integrating the data from vehicle-side and road monitoring devices, a more comprehensive and accurate environmental perception and information acquisition can be achieved. The Calibration of sensors, as an essential component of autonomous driving technology, has consistently drawn significant attention. Particularly in scenarios involving multiple sensors collaboratively perceiving and addressing localization challenges, the requirement for inter-sensor calibration becomes crucial. Recent years have witnessed the emergence of the concept of multi-end cooperation, where infrastructure captures and transmits surrounding environment information to vehicles, bolstering their perception capabilities while mitigating costs. However, this also poses technical complexities, underscoring the pressing need for diverse end calibration. Camera and LiDAR, the bedrock sensors in autonomous driving, exhibit expansive applicability. This paper comprehensively examines and analyzes the calibration of multi-end camera-LiDAR setups from vehicle, roadside, and vehicle-road cooperation perspectives, outlining their relevant applications and profound significance. Concluding with a summary, we present our future-oriented ideas and hypotheses.
\end{abstract}


\section{Introduction}

 \begin{figure}[ht]
 \centering
 \includegraphics[width=8.9cm]{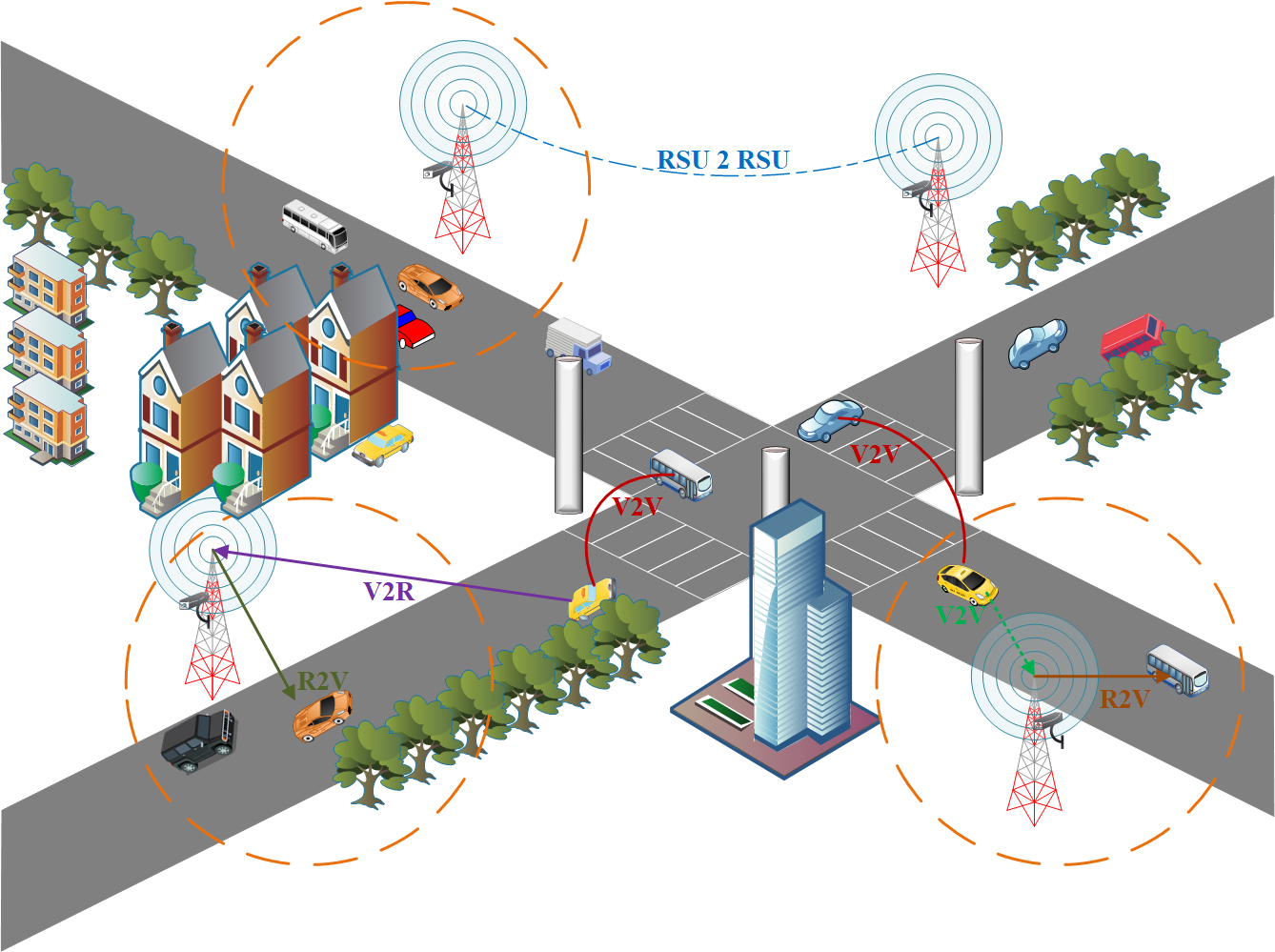}
 \caption{Schematic, intelligent connected cars and road infrastructure collaboration on a traffic scenario.} 
 \label{fig-zong1}
 \end{figure} 
 
 \IEEEPARstart{I}{n} a typical urban intersection scenario, effective data fusion between vehicle-side and road monitoring equipment is crucial. Whether on the vehicle side or within the road monitoring devices, both are equipped with visual and LiDAR sensors. By successfully integrating the data from vehicle-side and road monitoring devices, a more comprehensive and accurate environmental perception and information acquisition can be achieved. While LiDAR sensors provide high-precision distance and depth information, visual sensors capture and analyze visual information on the road, including traffic signs and the status of other vehicles. This comprehensive utilization of visual and LiDAR data from vehicle-side and road monitoring equipment is pivotal in ensuring the efficient operation and quality output of the traffic monitoring system. Through refined data fusion mechanisms, not only can the quality and efficiency of traffic management be enhanced, but a solid foundation can also be laid for the intelligent development of urban traffic systems. In busy urban intersections like these, effective fusion of vehicle-side and road data is a crucial element in ensuring traffic safety and flow. People have continuously increased the research investment in autonomous driving technology out of consideration for reducing traffic accidents\cite{Yeong_SSF}, saving energy and environmental protection in proper way\cite{Wadud_HH?} or liberating productivity\cite{Duarte_TIA}, etc. The autonomous driving technology people talk about often refers to the vehicle using sensors to collect information about its surroundings and real-time processing, to obtain a better understanding. And combined with location and mapping, path planning, decision making and vehicle control modules, the vehicle can be safely driven without being taken over\cite{Marti_ARS}. However, when performing the above functions, vehicle using multi-sensor systems often encounter the uncertainty of data fusion\cite{Wang_MFS} \cite{Xin_EKP}, that is, data need to be registered accurately through calibrated sensor parameters. Therefore, as the basis for these functions, the use and calibration of multi-sensor is particularly critical and fundamental. Camera and LiDAR, as the most commonly used sensor in autonomous driving\cite{Kocić_SSF}, is focused on this paper. The calibration of their peers and each other is the primary problem that engineers often face in the calibration of sensors for autonomous driving.
 
In addition, it should be noted that autonomous driving technology is gradually evolving from multi-sensor system at the single vehicle to multi-sensor fusion at multi-end cooperative, as shown in Fig. 1. At present, there are many concepts proposed, like vehicle to everything (V2X) which includes vehicle to vehicle (V2V), vehicle to infrastructure (V2I) and vehicle to pedestrian (V2P), and vehicle to network/cloud (V2N/V2C) network connections
. Thus, concepts like V2V or V2I in V2X using multi-sensor systems can be used to coordinate the vehicle with other ends to achieve a larger perception range, reduce blind zones and better detection effects\cite{Maruta_BVV}.  Therefore, only discussing the multi-sensor calibration of the single vehicle cannot well summarize the current research situation, in this paper, we start from the multi-sensor calibration of the vehicle, and also summarize the multi-end cooperative LiDAR-camera calibration related to the roadside and vehicle-road cooperation.

\begin{figure}[ht]
 \centering
 \includegraphics[width=8.9cm]{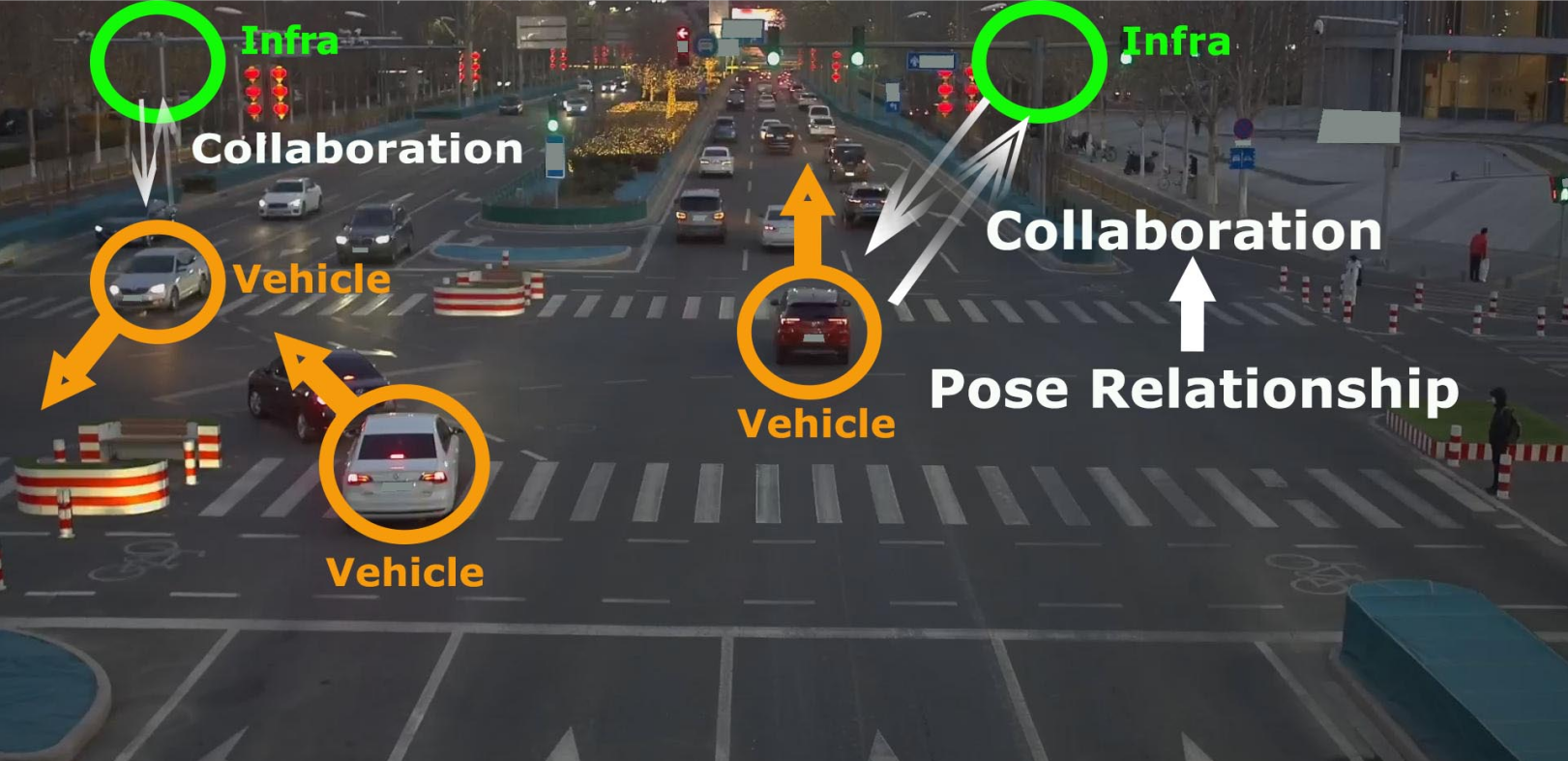}
 \caption{Schematic, intelligent connected cars and road infrastructure collaboration on a traffic scenario.Infra is short for road infrastructure.}
  \label{fig-1}
 \end{figure}

Thus, we could notice that calibration of camera-LiDAR, camera-camera and LiDAR-LiDAR is one of the essential problems in various aspects of autonomous driving technology. In perception problem, people often fuse camera and LiDAR in vehicle to achieve better accuracy in object detecting 
\IEEEpubidadjcol 
and tracking where its fusion of data to a same coordinate needs to obtain external transformation of sensors\cite{Zhao_F3L} \cite{Banerjee_OCL}. In addition, multiple sensors are calibrated in a multi-end collaboration, which has enhanced the perception range of the vehicle and reduced blind areas\cite{Craft_EVL},\cite{Song_ASC}. In localization problem, whether people use the positioning method of feature extraction and matching to obtain the correspondence of real-world dimensions in mapping\cite{Kim_RFE}, or the relative position relationship between multi-end sensors of multi-vehicle coordination (V2V)\cite{Song_ASC}
or vehicle-road coordination (V2I/V2R)\cite{Duan_VBE} for positioning, it is necessary to calibrate the external parameters of the sensors to obtain the relative position relationship, so as to register the data to obtain the joint map and position of vehicle. Thus, the calibration of multi-end cooperative camera-LiDAR is an essential problem for autonomous vehicle and provide basis for further application.

In this paper, the content will be separated into three parts of multi-sensor calibration with focusing on camera and LiDAR: Calibration between multiple sensors in a single end (vehicle and infrastructure), joint calibration between vehicle and infrastructure, and subsequent related tasks. The relationship between the three parts can be summarized as taking the vehicle as the main body and multi-end cooperative. The purpose is to enhance the perception or localization capabilities of the vehicle. The vehicle multi-sensor calibration has already had many research and applications. But the roadside, as another perspective of the infrastructure, cooperate with vehicle which can achieve better perception effect and cost savings\cite{Fleck_IAM}. And we also pay attention to the roadside itself multi-sensor calibration as the basis of vehicle road coordination. Vehicle-road cooperation \cite{Chen_IPS}, as a new concept, has more important significance in autonomous driving tasks. Through multi-end sensor calibration, the coordinate systems of the vehicle and the road end sensors can be accurately aligned to ensure that the information shared by them can be effectively fused under the same coordinate system, so as to achieve more comprehensive and accurate environmental perception and vehicle positioning.

The structure of the paper bellow is organized as follows: Section II explain the basic principles of camera-LiDAR, camera-camera and LiDAR-LiDAR calibration. Section III covers the calibration methods between multiple sensors in a single end (vehicle and infrastructure), and describes the commonality and difference between vehicle calibration and infrastructure calibration tasks. Section IV summarizes the joint sensor calibration between vehicle and infrastructure, and secction V summarizes the joint sensor calibration between vehicle and infrastructure, . Section VI introduces relative applications and significances. Section VII summarizes content and prospects for future.

\section{SENSOR CALIBRATION MATHEMATICAL PRINCIPLE}
The calibration of multi-sensor in multi-end we mentioned above mainly refers to the calibration of relative position transformation relationship between sensors, that is, the extrinsic calibration between sensors. Since our paper mainly focuses on camera and LiDAR, here we mainly discuss camera-LiDAR, camera-camera and LiDAR-LiDAR extrinsic calibration. Considering that the foundation of visual LiDAR calibration lies in the accurate calibration of camera intrinsic parameters and sensor synchronization, the paper will also provide a suitable introduction to the principles and representative works in these two aspects.

The intrinsic parameters and distortion coefficients of a camera represent its inherent characteristics and are typically set during the manufacturing process. Consequently, many manufacturers now provide the intrinsic parameters directly at the time of production. Moreover, the continuous enhancement in camera design has led to improved distortion control, resulting in a relatively slower pace of research and updates in intrinsic parameter calibration algorithms.However, it is undeniable that the intrinsic parameter calibration of the camera itself is an important prerequisite for the extrinsic parameter calibration between multiple sensors., which hence is briefly introduced here first, while the intrinsic of LiDAR are generally calibrated before leaving the factory and can be used directly. And the camera-LiDAR, camera-camera and LiDAR-LiDAR extrinsic calibration will be introduced later.
\subsection{Camera Intrinsic Calibration}
The intrinsic calibration of the camera refers to the relationship between the pixel coordinate system and the camera coordinate system, and usually people will do the de-distortion at the same time. \cite{Zhang_CC3} Noted that the intrinsic of camera are usually fixed before leaving the factory and we could get focal length f and pixel size (width k and height l) from the manufacturer. 

By referring Pinhole camera model, we can get the relation between a point in camera coordinate system P(X,Y,Z) and its projection in image coordinate system p(x,y) as.
\begin{equation}
x=\frac{f}{z}X,y=\frac{f}{z}Y
\end{equation}

The correspond position in discrete pixel coordinate system is (u,v). Let $f_x=\frac{f}{k},f_y=\frac{f}{l}$, and the relation between p(x,y) and (u,v) is
\begin{equation}
u=f_x\frac{X}{z}+c_x,v=f_y\frac{Y}{z}+c_y
\end{equation}

The intrinsic matrix is

\begin{equation}
\left[
\begin{array}{c}
     u\\
     v\\
     1
\end{array}
\right]
=
\frac{1}{2}
\left[
\begin{array}{ccc}
     f_x & o & c_x \\
     0 & f_y & c_y \\
     0 & 0 & 1
 \end{array}
 \right]
 \left[
 \begin{array}{c}
     X\\
     Y\\
     Z
 \end{array}
 \right]
\end{equation}
which can also be written as $Z_{P_{uv}} = KP$.

De-distortion operation is usually executed followed if this is a non-linear camera model.

Thus, K is the intrinsics matrix of camera. Camera intrinsic calibration mathematically means to get the matrix K here.

\subsection{Extrinsic Calibration}
In this paper, we mainly discuss the calibration between multi-end and multi-sensor, what needs to be obtained is the relative coordinate transformation relationship between sensors. As the LiDAR and camera that we concerned here have different properties, the calibration between them or with their peers often needs different methods. But in short, the relative coordinate transformation relationship is what we aim to obtain in the end. Therefore, we can summarize the extrinsic calibration between multiple sensors as follows.

\subsubsection{Camera-LiDAR Extrinsic}
\ 
\newline
\indent 
For camera-LiDAR extrinsic, the mathematical form is to compute the relationship between a homogeneous 3D point $P_i^L$ from point clouds and its homogeneous image projection $p_i^C$ \cite{Mishra_EC3}
, which is given by
\begin{equation}
p_{i}^{C}=K\left({ }^{C} R_{L} P_{i}^{L}+{ }^{C} T_{L}\right)
\end{equation}
where ${ }_{}^{C} R_{L}$ is rotation matrix parametrized by Euler angles $[\phi,\theta,\psi]^{T}$ and ${ }_{}^{C} T_{L}$  is the translation vector $ \left[t_{x}, t_{y}, t_{z}\right]^{T} $.

Thus, the camera-LiDAR extrinsic calibration is to compute the ${ }_{}^{C} R_{L}$ and ${ }_{}^{C} T_{L}$ which have total 12 parameters and need at least 4 correspond points pairs.

\subsubsection{Camera-camera Extrinsic}
\ 
\newline
\indent  
For camera-camera extrinsic, the mathematical form is to compute the relationship between two correspond points $p^{c_1}_i$ and $p^{c_2}_i$ in two cameras’ pixel, which is
\begin{equation}
    p^{c_1}_i = K_1(^{c_1}R_{c_2}K^{-1}_2+^{C_1}T_{c_2})
\end{equation}
where $^{C_1}R_{C_2}$ is rotation matrix and $^{c_1}T_{c_2}$ is translation vector.

We already know that all cameras in system have intrinsic matrix K, rotation matrix R and translation matrix T which transform points from the camera coordinate system to the fiducial coordinate system\cite{Pless_UMC}.

Actually, for multi-camera extrinsic calibration, people often transform the data from images into a common coordinate (cc). One method\cite{Brückner_IEA} is to use undistorted images points. Another one\cite{Zhang_MCL}is to use IMU coordinate. Thus, for two cameras
\begin{equation}
^{C_1}R_{c_2} = ^{c_1}R_{cc}\\^{c_2}R^{-1}_{cc}
\end{equation}
\begin{equation}
^{c_1}T_{c_2} = ^{c_1}T_{cc} + ^{c_1}R_{CC}\\^{C_2}T_{cc}^{-1} 
\end{equation}

Thus, the camera-camera extrinsic calibration is to compute the $^{c_1}R_{c_2}$ and $^{c_1}T_{c_2}$. And for multi-camera, we can also transform every image data from camera into a common coordinate. 
\subsubsection{LiDAR-LiDAR extrinsic}
For LiDAR-LiDAR extrinsic, the mathematical form is also to compute the relationship between two 3D point $P^{L_1}_i$ and $P^{L_2}_i$ from two LiDAR’s point clouds, which is
\begin{equation}
P^{L_1}_i = ^{L_1}R_{L_2}P^{L_2}_i + ^{L_1}T_{L_2}
\end{equation}

where $^{L_1}R_{L_2}$ is rotation matrix and $^{L_1}P_i$ is translation vector.

Thus, for two LiDARs
\begin{equation}
^{L_1}R_{L_2} = ^{L_1}R_{cc}\\^{L_2}R_{cc}^{-1}
\end{equation}
\begin{equation}
^{L_1}T_{L_2} = ^{L_1}T_{cc} + ^{L_1}R_{cc}\\^{L_2}T_{cc}^{-1}
\end{equation}
Thus, as extrinsic is relative position and the relative angle from the reference LiDAR sensors to the other\cite{Lee_ECM}
, the LiDAR-LiDAR extrinsic calibration is to compute the $^{L_1}R_{L_2}$ and $^{L_1}T_{L_2}$. And for multi-LiDAR, we can also transform every point cloud data from LiDAR into a common coordinate like a reference LiDAR.

Therefore, from the mathematical principle, we make it clear that the overall goal of our calibration is relative coordinate transformation, that is, pose transformation (position and angle). Next, for the calibration of mathematical parameters mentioned above, we will introduce different multi-sensor calibration methods about LiDAR and camera in three parts respectively: vehicle, roadside and vehicle-road cooperative.

\section{Fundamental Pillars}
The accurate calibration of visual LiDAR systems relies on two fundamental prerequisites: camera intrinsic calibration and sensor synchronization. Camera intrinsic calibration involves determining the internal parameters of the camera, such as focal length, principal point, and distortion coefficients, to ensure accurate conversion from pixel coordinates to world coordinates. Simultaneously, sensor synchronization ensures temporal alignment of data from different sensors, enabling effective data fusion and perception. In this section, we will focus on these two key prerequisites in visual LiDAR calibration: camera intrinsic calibration and sensor synchronization. We will review current mainstream methods for camera intrinsic calibration and sensor synchronization techniques.

\begin{figure*}[ht]
\centering
\includegraphics[width= \linewidth]{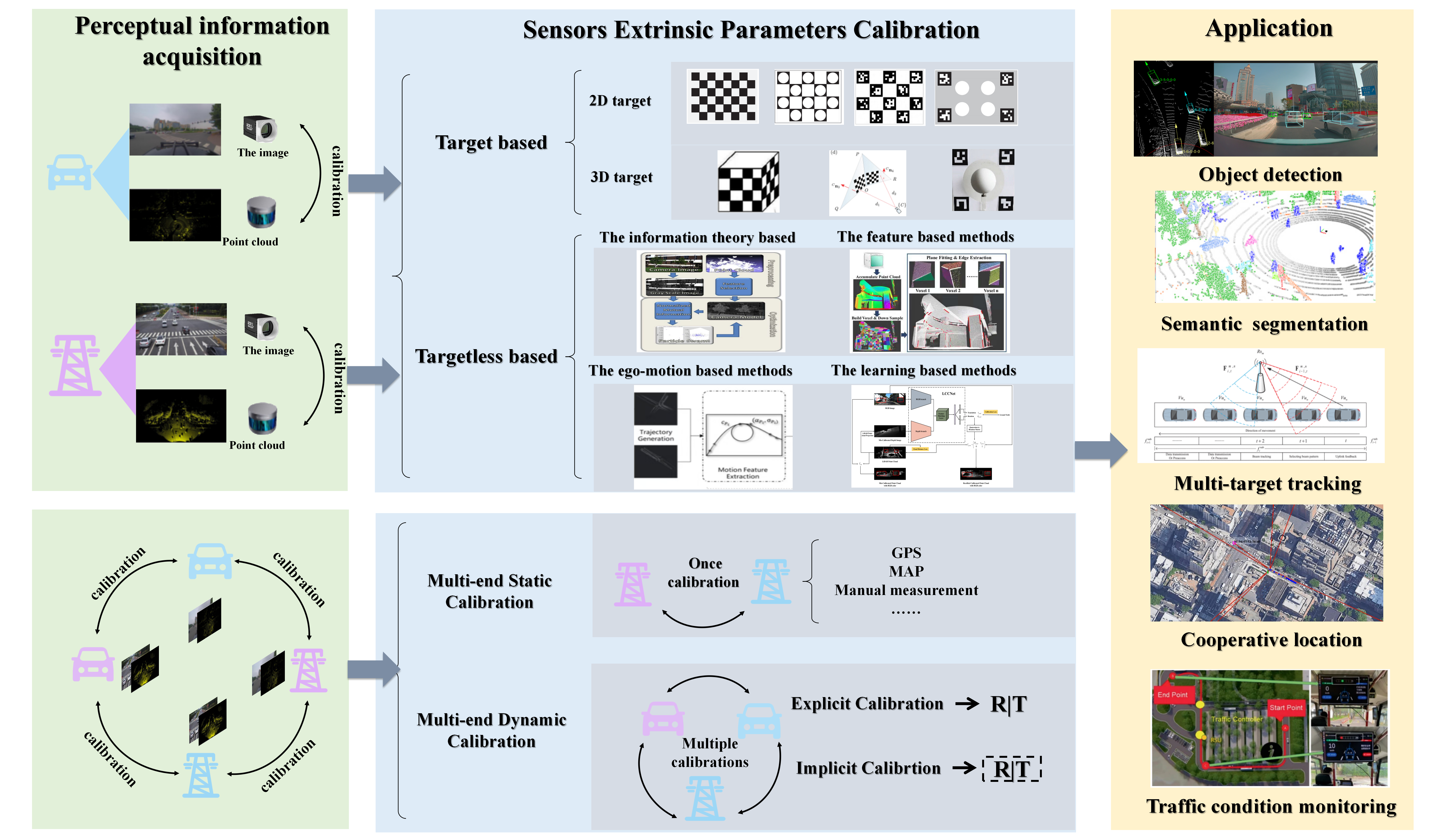}
\caption{Collaborative Visual-LiDAR extrinsic calibration process: (1)Data Collection for Calibration Objects: Camera and LiDAR data are collected simultaneously at the vehicle and infrastructure ends. Prior to this, the camera needs to undergo intrinsic calibration. (2)Individual Camera and LiDAR extrinsic calibration are separately completed at the vehicle and infrastructure ends to ensure consistent representation of camera and LiDAR data on the same end, thereby enabling single-end visual-LiDAR data fusion. The specific method varies based on the need for specific artificial markers, categorized into requiring target calibration and targetless calibration.(3) Collaborative extrinsic calibration is performed at the vehicle and road ends. Depending on whether specific extrinsic are directly output, methods are categorized into implicit and explicit methods. (4) The fused Visual-LiDAR data obtained from calibration can be used for various tasks after different decoding and post-processing, using different information in diverse applications.} 
\label{fig-zhu3}
 \end{figure*}

\subsection{Camera Intrinsic Calibration }
The accuracy of camera intrinsic calibration directly impacts the positioning and perception capabilities of visual LiDAR systems. By precisely estimating the camera's internal parameters, image distortions can be eliminated, and accurate mapping from pixel coordinates to world coordinates can be achieved, providing a reliable foundation for subsequent target detection, tracking, and 3D reconstruction. Research on camera intrinsic calibration techniques focuses on the point-to-point mapping relationship between the 3D spatial coordinates of objects and their 2D pixel coordinates in images. Common calibration methods typically use artificially designated calibration objects. These objects can be one-dimensional linear standard points\cite{Zhang_CCO}, two-dimensional checkerboard patterns\cite{Wu_Bo_AFM}, or three-dimensional checkerboard objects\cite{Zhai_DRT}, among others. Tsai\cite{Tsai_AVC} considered the camera's distortion factors and proposed a two-step method based on radial correction constraints, which effectively improved calibration accuracy, but the calibration process requires expensive equipment. Currently, the checkerboard calibration method proposed by Zhang Zhengyou\cite{Zhang_AFT} is quite popular. This method optimizes the camera's internal parameters through iterative calculations based on the different spatial positions of the checkerboard images and the pixel coordinates of the checkerboard corners in two-dimensional images, achieving intrinsic calibration. The precise extraction of checkerboard corner pixel coordinates is key to improving the accuracy of calibration results. To address the issue of low calibration accuracy due to highlights in the photographed pictures, Zhu Zhenmin et al.\cite{Zhu_CCM}proposed a camera calibration method based on optimal polarization information. This method obtains checkerboard correction images with the best polarization angle by adjusting the polarizer at different spatial positions, enabling precise extraction of checkerboard corners. Usenko\cite{Usenko_TDS} introduced a new dual-spherical camera model suitable for fisheye cameras, which, compared to models based on high-order polynomials, improves solution speed while maintaining original accuracy through spherical projection.
\subsection{Multisensor Synchronization}\label{AA}
Sensor synchronization ensures temporal consistency among data from different sensors, preventing data misalignment and error accumulation, thereby achieving accurate multisensor data fusion and collaborative operation. Synchronization methods are generally divided into hardware synchronization and software-based synchronization. Hardware synchronization involves using sensors that support hardware triggering for triggered synchronization, i.e., using hardware triggers to control the sensor's frame rate directly through physical signals. GNSS, an essential sensor in autonomous driving, comes with a second pulse generator and its clock is corrected by satellite atomic clocks, typically achieving an accuracy of 10 ns. It serves as a true value calibration for the synchronization controller, providing timing to the entire system while performing its positioning function\cite{Li_ARM}. Marsel Faizullin and others\cite{Faizullin_OIT} proposed a microcontroller-based platform for simulating GPS signals, extendable to any number of sensors beyond LiDAR-IMU. Hannes Sommer and others\cite{Sommer_ALS} used simple external devices, including LEDs (for cameras) and photodiodes (for LiDAR), to achieve time calibration, trading off the cost of additional equipment for high-speed and high-precision time offset estimation. All the above methods require a hardware-triggered interface or support from other hardware, limiting their use in all scenarios. In contrast, software-based methods have become increasingly popular due to their lower hardware requirements. The ROS-based message filtering method ensures that messages published by multiple publishers are synchronized in time for processing in callback functions, performing poorly in high-precision data fusion scenarios. Wang and others\cite{Wang_TSO} proposed a posture estimation model and environmental line feature extraction method to address the time delay issue, achieving real-time online calibration of cameras and LiDAR. Kodaira and others\cite{Kodaira_SSP} introduced an online joint spatiotemporal calibration algorithm for vision-LiDAR, using visual odometry to estimate the time delay between sensors.
\section{ SINGLE-END CAMERA-LIDAR CALIBRATION}
In this section, we focus on the multi-sensor calibration method at the the single end between roadside and vehicle which mainly involve camera and LiDAR and discuss its significance and difference. The overview diagram is shown in Figure~\ref{fig-zhu3}.

Calibration tasks refer to the calibration of equipments to accurately determine the relative positional and externally-referenced relationships between them in an autonomous or intelligent driving system. Such a calibration task is to ensure accurate alignment of multiple sensors on the same carrier for effective perception and environment understanding, and to provide accurate inputs for autonomous vehicle navigation and decision making.

In automatic driving technology, most of the existing sensor calibration schemes are applicable to the vehicle since the vehicle is the scenario for mainstream calibration tasks. However, as the roadside is a fixed end in the driving scene, we cannot deny that it can play an irreplaceable  role in autonomous driving.

On traditional traffic roads, pole-shaped roadside infrastructure is often installed at the roadside. People usually install cameras, speed radar, or hang traffic signs on it, and there are also electronically displayed traffic lights at the intersection. This is to assist traffic management and help the network of driving vehicles understand road information.

In autonomous driving technology, roadside infrastructure also has a similar irreplaceable role. The Intelligent Roadside Unit (IRSU) is one of concepts in Collaborative Intelligent Transportation Systems which is proposed to be used as an equipment for cooperation between vehicle and roadside\cite{Shan_DCP}. Thus, this kind of roadside mentioned here, during autonomous driving, can provide more diverse, wider range and more accurate information for the vehicle. Moreover, IRSU can also bring more intelligent information to upper traffic management.

As vehicle often encounter the limitations of the perception range, the existence of perception blind zones\cite{Xiang_MFA} and the high cost of vehicle sensor system\cite{Varghese_OAV}, the participation of the roadside provides a new method to solve these. People often use the sensor system in roadside to map the road in a wider range\cite{Lv_LCI} or cooperate with vehicle sensors to do localization\cite{Duan_V2I}, or target detection of road targets\cite{Xiang_MFA}
, which can transfer more information to the vehicle. And if the roadside sensor system has met certain perception capabilities, through communication and other ways to serve multiple vehicles at one time, the vehicle end can reduce the appropriate number of sensors or reduce the requirements of the upper computer, to save the cost of single vehicle.

Thus, we can realize that multi-sensor calibration at the roadside is essential, which not only is the basic of roadside multi-sensor system in terms of the assistance and the cooperation with vehicle, but also provide more different inspirations in perception solutions. 

From a practical point of view, vehicle calibration focuses more on the application in real driving vehicles, which needs to consider the dynamic characteristics of the vehicle and the driving environment, while road test equipment calibration pays more attention to the accuracy and stability of the test equipment to ensure the reliability and validity of the experiment.That’s to say, vehicle calibration usually requires real-time calibration to adapt to changes in the vehicle's attitude, while road test equipment calibration can be carried out before the experiment for a more stable calibration.

For single-end camera-LiDAR calibration, diverse and comprehensive solutions are available to cater to different requirements. These solutions encompass both target-based and targetless-based methods. In contrast, research into road test equipment calibration is still in its nascent stages, with only a limited number of calibration strategies designed explicitly for road test equipment. Often, existing vehicle calibration schemes are either adapted for roadside usage or generalized approaches without a strong emphasis on sensor platform are extended to roadside scenarios, subsequently optimized and adjusted according to the characteristics of the environment. Consequently, this chapter discusses calibration strategies for both vehicle and road test equipment calibration.

Based on whether the calibration process needs targets, the calibration methods described as follows can be broadly divided into two categories: target-based methods and target-less methods. As mentioned before, we only talk about two types of sensors here: LiDAR and camera.

\subsection{Target-based Method }\label{AA}

According to the dimensional characteristics of markers, the methods are divided into 2D based calibration method and 3D based calibration method.
he target-based method mainly includes 1D calibration object, 2D calibration object and 3D calibration object. They have their own characteristics and are suitable for calibration of different multi-sensor combinations. Table I summarizes the research status of single-end camera-LiDAR calibration methods using targets. Table I summarizes the research status of single-ended camera-LiDAR target calibration methods

\begin{figure}[ht]
 \centering
 \includegraphics[width=8.9cm]{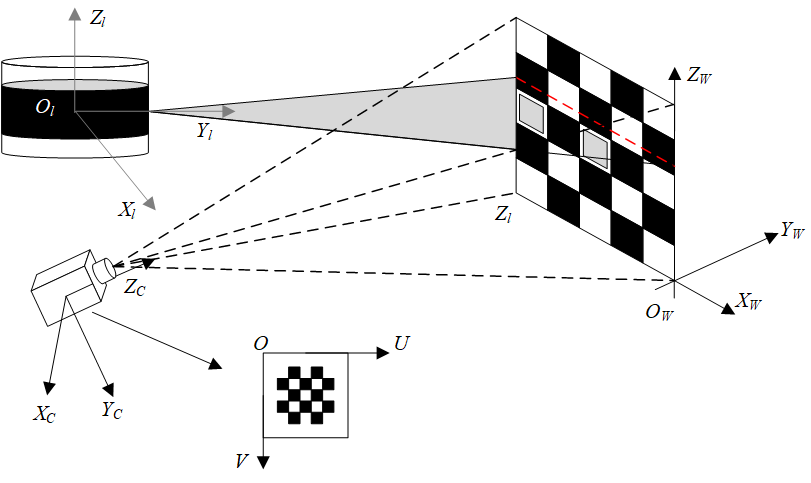}
 \caption{Illustration of Visual-LiDAR Calibration Using a 2D Checkerboard Calibration Board.}
  \label{fig-target-based}
 \end{figure}

\begin{table*}
\caption{SUMMARY OF TARGET-BASED METHODS.}
\label{tab:my-table}
\renewcommand{\arraystretch}{1.3} 
\setlength{\tabcolsep}{3pt} 
\begin{tabularx}{\textwidth}{|c|c|c|>{\centering\arraybackslash}X|>{\centering\arraybackslash}X|>{\hsize=1.1\hsize\centering\arraybackslash}X|>{\hsize=0.9\hsize\centering\arraybackslash}X|>{\hsize=0.8\hsize\centering\arraybackslash}X|>{\centering\arraybackslash}X|}
\hline
\multicolumn{1}{|c|}{\textbf{}} & \multicolumn{1}{c|}{\textbf{Paper}} & \textbf{Year} & \textbf{Sensor assembly} & \textbf{Targets} & \multicolumn{2}{c|}{\textbf{Characteristics}} & \textbf{Matching strategy} & \textbf{Open source or not} \\
\hline
\multirow{6}[10]{*}[1.5ex]{\begin{tabular}[c]{@{}c@{}}2D\\Object\end{tabular}} & \cite{AECM} & 2022 & multicamera-multiLIDAR & Perforated 2D plate & \multicolumn{2}{c|}{ArUco marker + Geometrical features}  & Umeyama algorithm for solving the least squares problem & Yes\\
\cline{2-9}                           
& \cite{ACML} & 2018 & multicamera-multiLIDAR & Rectangular cardboard and ArUco markings & \multicolumn{2}{c|}{Select manually} & Bundle Adjustment & No \\
\cline{2-9}
& \cite{AAL} & 2013 & single camera-single LIDAR(2D) & Calibration plate & \multicolumn{2}{c|}{Straight line + Depth discontinuity}  & RANSAC & No\\
\cline{2-9}
& \cite{AECL} & 2017 & multicamera-single LIDAR & On a flat target & \multicolumn{2}{c|}{Image edge + Depth information} & Iterative Closest Points & Yes\\
\cline{2-9}
& \cite{OCC} & 2017 & single camera-single LIDAR & Checkerboard pattern & \multicolumn{2}{c|}{Range conversion + Deep discontinuity filter point}  & Least square method + Levenberg-Marquardt & No\\
\cline{2-9}
& \cite{AECA} & 2021 & single camera-single LIDAR & Checkerboard pattern & \multicolumn{2}{c|}{The center of the checkerboard and the normal vector} & Genetic Algorithm & No\\
\hline
\multirow{5}[10]{*}[1.5ex]{\begin{tabular}[c]{@{}c@{}}3D\\Object\end{tabular}} & \cite{PAC} & 2017 & single camera-single LIDAR & Case & \multicolumn{2}{c|}{Plane of case} & RANSAC+ PnP & No\\
\cline{2-9}
& \cite{ANC} & 2020 & single camera-single LIDAR & V-shaped object & \multicolumn{2}{c|}{V-break} & Nonlinear least square method & No\\
\cline{2-9}
& \cite{ANM} & 2018 & single camera-single LIDAR(2D) & Convex V structure & \multicolumn{2}{c|}{Triangular plate + Laser point} & Levenberg-Marquardt & No\\
\cline{2-9}
& \cite{ACMC} & 2018 & multicamera-multiLIDAR & Sphere & \multicolumn{2}{c|}{Geometric feature} & Numerical solver & No\\
\cline{2-9}
& \cite{EC2} & 2015 & single camera-single LIDAR(2D) & Orthogonal trihedron & \multicolumn{2}{c|}{Normals of three vertical planes} & RANSAC & No\\
\hline
\end{tabularx}
\end{table*}

\subsubsection{2D Calibration Object}
2D calibration object mentioned here often refers to the 2 D calibration board. Because the calibration board is more convenient to use, has high accuracy, and generally does not appear self-occlusion, it is widely used in calibration tasks, but can only be applied to non-large field of view calibration. Therefore, people generally uses calibration boards for non-large field of view calibrations such as multi-sensor calibrations at the single road end or at the vehicle end. There is a considerable number of 2D calibration methods available\cite{L.Yaopeng R3L}, mainly aimed at enhancing accuracy. However, given the focus of this paper on collaborative calibration between vehicle and road-side sensors, we will refrain from delving extensively into two-dimensional calibration approaches to avoid overburdening the content with unnecessary details.

The researchers used 2D markers with different shape differences for the study. Jorge Beltran et al. describe an automated external calibration method in which 2d orifice plates are used as calibrators during the study to precisely calibrate external parameters involving LIDAR-camera sensor pairs. \cite{AECM} Zoltan Pusztai et al. studied and introduced a calibration method using rectangular cardboard and ArUco markers as calibration objects.
a new LiDAR-camera calibration is introduced, which uses ordinary cardboard boxes. It achieves high accuracy and can be used with multiple sensors. A new Bundle Adjustment (BA)-based technique is introduced to reduce the overall error of LiDAR-camera system calibration. The method is evaluated on both synthetic and real-world data, competing against state-of-the-art techniques.Moreover, a technique is also presented for estimating the car body border, as a 2D bounding box, with respect to the calibrated sensors\cite{ACML}. Chao X. Guo et al introduce a simple, yet powerful, calibration procedure, where the calibration target used only contains a straight line. Moreover, we do not require measuring the laser intensity which makes our method applicable to a wider range of LIDARs. They formulate the LIDAR-camera extrinsic calibration problem as a least-squares minimization problem and solve it analytically to find the optimal values for the 6 dof unknown transformation\cite{AAL}. Carlos Guindel et al. developed an automated calibration process that enables accurate calibration by utilizing circular holes, a feature common to liDAR and the camera's field of view\cite{AECL}.  Bianca-Cerasela-Zelia Blaga et al. used a checkerboard as a marker to accurately align and synchronize the two devices by extracting common features in camera images and liDAR scan data\cite{OCC}.  Surabhi Verma et al., using a checkerboard calibration plate as a calibration object, describe a method for online cross-calibration of cameras and LIDAR designed to automatically adjust calibration parameters between the two sensors. Errors due to sensor position drift can be corrected in real time\cite{AECA}. 
\subsubsection{3D Calibration Object}
3D calibration object has a unique advantage in the calibration task between multi-end sensors which its 3D nature allows it to not care about its own orientation. The only thing is to make sure that it is in the common field of view of the multi-end sensor. Besides, symmetrically shaped object is often used so that the data perceived by multiple angle sensors is similar. Thus, the 3D calibration object is very suitable for the calibration of multiple roadside sensors in the large field of view.Common 3D calibration objects include spheres, V-shaped objects and polyhedra.
Julius Kümmerle et al presents a novel automated calibration method that utilizes sphere characteristics to achieve precise LiDAR-camera alignment\cite{ACMC}. In Shoubin Chen et al's study, V-shaped breakpoints were used as calibrators, combined with numerical optimization strategies of nonlinear least squares to improve the accuracy and efficiency of calibration. \cite{ANC} Wenbo Dong et al used the point-to-plane constraint of a single observation of a V-shaped object to uniquely determine the relative position between two sensors.\cite{ANM} Ruben Gomez-Ojeda et al. developed a method that uses specific geometric features (such as corner points) in three-dimensional environments observed by cameras and lidar to determine the relative position and orientation between these two types of sensors through mathematical models and algorithms.\cite{EC2} Pusztai et al uses a box with three vertical edges as the calibration object. By extracting the edges of the box, the 3D vertices of the box are estimated and matched with the 2D vertices extracted from the image.\cite{PAC} 

To sum up, both 2d and 3d calibration are unique advantages and application scenarios in the calibration process of multi-sensor system, which shows the importance of ensuring accuracy and efficiency.

\subsection{Targetless-based Method}
The targetless-based method usually focuses on the natural features from real scenarios and calibrate based on a variety of high-dimensional attributes or features, which concede some of the accuracy and generalizability to acquire the advantage of automatic calibration.. The feature extracting and matching is often the core of methods. For calibration between multiple sensors on a single roadside and between sensors on multiple roadsides, and even for single sensors on a single roadside, the method is slightly different due to different viewing angles and aims. According to the survey from Li X et al., The automatic targetless LiDAR-camera calibration methods can be divided into information theory based methods, feature based methods, ego-motion based methods and learning based methods\cite{taylor2013automatic}. Most of these methods can be generalized to a variety of scenarios, with the one exception that the ego-motion based methods can only be utilized at the vehicle end due to the limitation of its mechanism.This section categorizes currently available targetless-based methods based on single-end and multi-end. Single-end comprises vehicle and single roadside.

The multi-sensor calibration problem of single-end is relatively simple due to the relative stationarity, close distance and similar angle of view between multi-sensor. Thus, many single-end’s multi-sensor calibration methods are applicable which mostly are about camera-LiDAR.  
\begin{table*}[t]
\caption{SUMMARY OF TARGETLESS-BASED METHODS.} 
\label{tab:my-table}
\begin{tabular}{|c|c|c|c|p{2.5cm}|p{3cm}|p{3cm}|} 
\hline
 & \textbf{Paper} & \textbf{Year} & \textbf{Author} & \textbf{Sensors} & \textbf{Features} & \textbf{Methods} \\
\hline
Information Theory Based & \cite{taylor2013automatic} & 2021 & Taylor Z et al. & Camera-LiDAR & LiDAR/Camera data & Umeyama algorithm \\
 & \cite{koide2023general} & 2023 & Koide K et al. & Camera-LiDAR & 2D-3D correspondences & Normalized information distance \\
\hline
Feature-Based Methods & \cite{zhang2021line} & 2021 & Zhang X et al. & Camera-LiDAR & Lines & Alignment \\
 & \cite{yuan2021pixel} & 2021 & Yuan C et al. & Camera-LiDAR & Edges & Alignment \\
 & \cite{wang2021roadside} & 2021 & Wang L et al. & Camera-LiDAR & Trajectories & Matching \\
 & \cite{Annkathrin2019Providentia} & 2022 & Krämmer A et al. & Multi-Camera/RADAR & Points & Alignment \\
 & \cite{muller2019laci} & 2019 & Müller J et al. & Stereo camera/Laserscanners & Vehicle's position/pose & Matching \\
 & \cite{10054792} & 2022 & Jing X et al. & Camera-LiDAR & 2D/3D points & Alignment \\
\hline
Ego-Motion Based & \cite{OCC} & 2017 & Nedevschi S et al. & Camera-LiDAR & Edges & Cross calibration \\
 & \cite{schneider2017regnet} & 2017 & Schneider N et al. & Multimodal sensors & Distinct features & Deep Neural Networks \\
\hline
Learning Based & \cite{chien2016visual} & 2016 & Chien H J et al. & Camera-LiDAR & SURF & Visual odometry \\
 & \cite{2020SOIC} & 2020 & Wang W et al. & Camera-LiDAR & Semantic information & Minimize cost function \\
\hline
\end{tabular}
\end{table*}

\subsubsection{The information theory based methods.}
Some researchers are also investigating mutual information-based calibration methods, Taylor tackled the constraints of target-based calibration methodologies by introducing a method reliant on mutual information between sensors for extrinsic parameter calibration\cite{taylor2013automatic}. This approach stands out for its enhanced robustness, eliminating the necessity for dedicated calibration targets. Building upon this concept, Kenji Koide and colleagues extended the idea by incorporating Normalized Information Distance (NID) to enhance calibration accuracy\cite{koide2023general}. This approach carries significant advantages alongside certain limitations. One of its strengths lies in mutual information being an information metric that doesn't rely on specific calibration objects, enabling calibration across different environments and scenarios. However, mutual information-based calibration methods also come with some drawbacks. Firstly, they demand high-quality sensor data with precise data matching and geometric correspondence. Moreover, due to the computational complexity of mutual information calculations, this method may be time-consuming for processing and optimization on large datasets. Additionally, the effectiveness of mutual information methods can be affected by factors such as sensor noise, occlusions, and dynamic environmental changes, leading to a potential decline in calibration accuracy. Lastly, this method may exhibit suboptimal performance in extreme scenarios, like situations with substantial sensor view differences or high levels of occlusion.
\subsubsection{The feature based methods}
The vehicle end and road end calibration methods will be different in the selection of features to meet different calibration requirements and environmental conditions. So this section is done separately for the two types of schemes.

As to the vehicle method, Zhang et. al proposed a calibration method that does not depend on any specific scene\cite{zhang2021line}. They uses edge features in the scene to complete the matching between the two types of data, but due to the complex traffic environment and many edge features, it is difficult to complete effective alignment\cite{yuan2021pixel}. The linear features was extracted \cite{zhang2021line}, using a projection method to conduct adaptive optimization. Since the real world has more linear features and clutter, if it does not consider how to filter line features, the method is prone to lower computational efficiency of the algorithm because of poor linear feature extraction. 

Regarding the single roadside method, Wang L et al. used an intelligent methodology where the camera and radar installed in roadside pole do object detection respectively, and then the trajectory of the detected object under the camera and radar frame is obtained respectively, and finally the trajectory matching is performed for calibration[45]. Then We can also obtain the object pair in matched trajectories. The aim of this auto-calibration method is to find a relationship between radar coordinates and video coordinates. Krämmer A et al. developed another method for the multiple camera and radar system in roadside involving vanishing points\cite{krammer2019providentia}. They firstly manually measure the relative translation of all sensors and reference point to obtain approximate initial parameters of extrinsic calibration algorithms, and then refine the camera poses using vanishing point methods\cite{kanhere2010taxonomy} that utilize parallel road markings to find vanishing points. Finally, they manually refine the alignment of the measurement points with each other. Müller J et al. proposed an algorithm for infrastructure sensors calibration\cite{muller2019laci}. The algorithm matches cooperative intelligent vehicle (CIV) detected in sensor frame with position/pose information from vehicle via Cooperative Awareness Messages (CAM)\cite{etsi2011intelligent} and applied to the calibration of multi-sensor traffic monitoring system\cite{buchholz2018digital} consisting of four laserscanner and one stereo camera. Besides, Jing X et al. developed a feature extracting and matching method to calibrate roadside camera and LiDAR\cite{zhang2021line}. The 2D features and 3D features are firstly extracted from color images and point cloud respectively. And then the distance transform image is obtained from a 2D feature map and used as a reprojection error energy function for optimization.

In short, In vehicle calibration, the selection of features often takes into account the dynamic nature of the vehicle and the variability of sensor data. For instance, features such as the vehicle's contour, lights, and license plate can be chosen as calibration targets to accommodate the diverse vehicle poses and environmental conditions. Furthermore, due to potential vehicle vibrations and perturbations during motion, feature selection needs to exhibit a certain level of robustness. Conversely, in roadside calibration, sensors are typically fixed in specific positions, leading to a greater emphasis on static scene features in feature selection. For instance, fixed elements like road markings, traffic signs, and road edges can be chosen as calibration objects, as they tend to remain relatively stable over extended periods, making them suitable for roadside calibration. Moreover, given the fixed installation positions of road-side sensors, feature selection may be more tailored to the specific environmental conditions of those positions.

\subsubsection{The ego-motion based methods}
In vehicle calibration, the vehicle's motion state is estimated using data from inertial measurement units (IMUs) during its movement. Subsequently, calibration is achieved by registering camera and LiDAR data with the estimated vehicle motion state. 

Nedevschi S. et al. employed feature matching to identify sensor offset while the vehicle was in motion\cite{nedevschi2017online}. A different approach utilized odometry data to estimate the camera-LiDAR motion relationship\cite{chien2016visual}. Furthermore, N. Schneider. et al. utilized front and rear camera images to reconstruct urban scenes in 3D and assess attitude changes, while also performing ICP calculations on LiDAR point clouds to capture spatial transformations.

However, in road-side calibration, the application scope of motion-based methods is relatively limited. This is due to the fact that the positions of most road-side sensing devices are fixed, preventing direct access to sensor motion information.
\subsubsection{The learning based methods}
Recent years have witnessed rapid advancements in machine learning techniques across various domains. Notably, the utilization of machine learning approaches for extrinsic calibration of visual LiDAR data has garnered significant attention \cite{ye2021keypoint}. These methods excel at capturing intricate non-linear mappings between sensor modalities. Nevertheless, these techniques face limitations, including the challenge of ensuring the adaptability of calibration models to diverse and novel scenes, given that training data might not encompass all possible scenarios\cite{iyer2018calibnet}, \cite{lv2021lccnet}, \cite{yuan2020rggnet}, \cite{sun2022atop}. Moreover, deep learning models' performance can be sensitive to the quality and quantity of training data, curtailing their suitability for real-world scenarios marked by complex data collection constraints. This scarcity of labeled training samples can detrimentally impact calibration outcomes.

Some scholars have also proposed schemes to use semantic information. The utilization of semantic information in sensor calibration involves leveraging static environmental cues, such as road structures and stationary traffic elements, to achieve calibration accuracy. This approach capitalizes on the consistency of these elements across various scenarios, ensuring reliable calibration results. By harnessing the inherent stability of semantic cues, this method offers real-time calibration feasibility and reduces the need for complex, dynamic calibration setups. However, while semantic information remains relatively stable, challenges include potential error accumulation in complex traffic scenarios and a dependence on accurate semantic segmentation for optimal performance.\cite{ye2021keypoint} proposed a semantic information-based extrinsic online calibration method. They introduced the method of semantic particle, using cars and pedestrians as marker to construct the PnP paradigm. The approach effectively extracts effective features from messy real-world data. However, cars and pedestrians are not stable markers in autonomous driving scenarios, such as neither pedestrians nor cars shall appear on an empty highway. The high-speed moving cars appearing in the LiDAR may cause point cloud distortion. The point cloud distortion caused by dynamic targets and the unstable distribution of vehicles and pedestrians in the traffic scene affect the calibration accuracy.  The semantic information is relatively stable in static environment, but the accuracy still needs to be improved in highly dynamic traffic scenarios, and the error may accumulate, especially in complex traffic situations, the error may increase rapidly. And it is very dependent on the quality of semantic segmentation: the accuracy and stability of the calibration method are affected by the quality of semantic segmentation, and inaccurate semantic segmentation may lead to unstable calibration results. Therefore, this method needs to focus on how to obtain stable semantic features in highly dynamic traffic scenes, and avoid using semantic information of dynamic objects that will bring unknown deviations.
\subsection{Conclusion}
To summarize, for the different needs of single-end multiple camera-LiDAR calibration, people's solutions are relatively various and complete where both target-based methods and targetless-based methods are involved. 

In the context of vehicle LiDAR-camera calibration, the target-based methods, while effective, exhibit limitations in addressing sensor offsets that may arise during vehicle motion. These methods are better suited for scenarios such as factory calibration or periodic recalibration, where dynamic environmental conditions are controlled. However, they may struggle to mitigate issues arising from real-time deviations in sensor alignment during vehicle operation. The targetless-based methods have emerged as a solution to the challenge of dynamic objects affecting calibration in traffic scenes, as well as the inherent instability of calibration results in complex traffic environments. They offer the advantage of adaptability to mobile surroundings, yet also present challenges. Avoiding the impact of dynamic objects requires sophisticated methods to filter out their influence on calibration accuracy. Additionally, the lack of universally recognized evaluation tools and comprehensive testing datasets tailored to the degree of sensor offsets poses an ongoing challenge that demands attention.

In the context of roadside LiDAR-camera calibration, for the target-based method, we can see that different target objects correspond to different demand scenarios, 1D calibration objects are more suitable for calibration of multi-camera on multi-end and multi-view in large scenes, 2D calibration objects which often are calibration boards are more suitable for camera-LiDAR calibration at the single-end, and 3D calibration objects are more suitable for multi-end LiDAR calibration in large scenes. It can be found here that due to the installation position and angle of roadside sensor are fixed and not easy to change which leads to the relative low demand for real-time calibration, there are still many methods that adopt calibration objects. However, due to the bird’s eye view angle and high installation height, this kind of calibration is different from that in the vehicle in sensor’s view angle and range.

For the targetless-based method, we can see that calibration of different types of sensors is often more difficult between multi-end with large differences in view angles. For calibration between multi-sensor on single-end, people can more focus on the camera-LiDAR calibration which involves matching vehicle trajectories using object detection or feature extracting and matching, etc. Table II summarizes the research status of vehicle camera-LiDAR and roadside camera-LiDAR calibration targetless methods.

These efforts aim to bridge the gap between theoretical advancements and practical implementation, enhancing the accuracy and robustness of sensor calibration techniques in the complex environment of Intelligent and connected vehicle In addition, we also noticed that in some roadside multi-sensor system methods, if it is a multi-sensor system on single roadside, people sometimes directly use the fixed extrinsic value provided \cite{wang2021cooperative}, \cite{dai2022roadside}. The values were obtained by means of manual measurements. and if it is a multiple roadside sensor, people sometimes use GPS to directly get relative positions of sensors\cite{wang2021cooperative}. However, the position accuracy is difficult to guarantee, and the Angle information between sensors cannot be obtained.
\section{Multi-end sensor calibration}
\label{multiendCalibration}

In modern intelligent systems, multi-node sensor perception technology plays a crucial role. By deploying sensors at various endpoints such as vehicles, infrastructure, and mobile devices, comprehensive environmental information is captured and real-time data is integrated for processing. This extensive perceptual capability not only enhances the system's understanding of and responsiveness to the external world but also significantly improves the accuracy and efficiency of decision-making.

To achieve this efficient collaboration of multi-node sensors, precise sensor calibration is essential. Multi-node sensor calibration is a key step in ensuring data accuracy and perceptual consistency, directly impacting the quality of data fusion and the overall performance of the system. This section will discuss this critical issue in detail.

Currently, multi-node calibration technology is in a phase of rapid development, particularly demonstrating its importance in applications such as autonomous vehicles, robotic navigation, and intelligent transportation systems. However, the practice and development of multi-node calibration technology face multiple challenges. Firstly, the complexity of scenarios directly increases the complexity of algorithms. For example, in urban environments, complex traffic conditions, diverse architectural layouts, and various static and dynamic obstacles require sensors to accurately identify and adapt. Secondly, in multi-node sensor systems, especially those with sensors that are far apart, the common viewing area may be very limited, which restricts the efficiency and accuracy of calibration. Additionally, high-quality feature extraction is crucial for cross-sensor data correlation, demanding that algorithms maintain feature stability under various conditions. The rapidity of environmental changes, such as lighting and climate variations, also requires calibration methods to be highly adaptable and flexible to adjust and optimize calibration parameters in real-time. Furthermore, multi-node calibration technology is closely related to data transmission rates and timing synchronization. Delays or instability in data transmission and errors in timing synchronization can significantly impact calibration accuracy.

In summary, the development of multi-node calibration technology not only requires efficient and accurate algorithms but also needs to address multiple challenges posed by external environments and operating conditions.

In this chapter, we categorize multi-node calibration into two main types: multi-node static calibration and multi-node dynamic calibration. These will be elaborated on in the following sections.

\begin{table*}[htbp]
\centering
\caption{SUMMARY OF MULTI-END STATIC CALIBRATION METHODS.}
\begin{tabularx}{\textwidth}{@{}| >{\hsize=.5\hsize\centering\arraybackslash}X| >{\hsize=.2\hsize\centering\arraybackslash}X| >{\hsize=.7\hsize\centering\arraybackslash}X| >{\hsize=.8\hsize\centering\arraybackslash}X| >{\hsize=1.8\hsize\centering\arraybackslash}X| @{}}
\hline
\textbf{Method} & \textbf{Year} & \textbf{Sensor type \& Scenario} & \textbf{Feature type} & \textbf{Matching \& Optimization} \\
\hline
TRR\cite{wu2019points} & 2019 & LiDAR(Inf) + LiDAR(Inf) & point feature (manual selected) & triangle matching + ground surface adjustment \\
\hline
ICNSC\cite{zhang20203d} & 2020 & LiDAR(Inf) + LiDAR(Inf) & point feature (3d target) & self-designed robust ICP \\
\hline
OLT\cite{lv2019revolution} & 2019 & LiDAR(Inf) + LiDAR(Inf) & point feature (ground point) & rotation and revolution-based ground surface matching \\
\hline
TITS\cite{ren2023trajmatch} & 2023 & LiDAR(Inf) + LiDAR(Inf) & trajectory feature + semantic feature & semantic feature matching + trajectory-level matching \\
\hline
ITSC\cite{zhao2023spatial} & 2023 & Camera(inf) + Camera(inf) & motion feature & monocular localization method based on deep learning \\
\hline
\end{tabularx}
\label{tab:multi-end-static-calibration}
\end{table*}

\subsection{Multi-end Static Calibration.}

Multi-node static calibration primarily involves the calibration of sensors between infrastructures, where the sensors are relatively fixed in position, such as cameras on traffic lights and radars along roadsides. The focus of this type of calibration is on how to extend single-node sensor calibration to scenarios with a wide field of view and limited common viewing areas, ensuring that each sensor can provide data stably and accurately.

Although many existing single-node calibration methods are theoretically applicable to multi-node static calibration scenarios, their effectiveness and feasibility in practical multi-node applications have not yet been widely verified through experiments and testing. Additionally, since the positions of sensors in static calibration are relatively fixed, manual operation is somewhat permissible, which can reduce the requirements for automation and ease the difficulty of the task in practical applications. This can be seen as an advantage, but on the other hand, it also increases the cost of manual maintenance.

Currently, research on multi-node collaborative calibration is still in its early stages, with most studies focusing on multi-node LiDAR systems. Moreover, these studies are primarily based on proprietary datasets, lacking a recognized, unified standard to evaluate and compare the effectiveness of different multi-node calibration methods. This not only limits further technological development but also affects the practical application and widespread adoption of these technologies.

The study \cite{zhang20203d} utilizes a 3D calibration object covered with special retroreflective materials for multiple roadside LiDAR systems to address the point cloud data alignment issues in complex traffic environments. These retroreflective materials enable the LiDAR to capture points on the reference system with high reflectivity, simplifying the process of feature point identification and extraction. An intensity-based filtering method is used to distinguish the points on the calibration object from other low-reflectivity points, and a density-based spatial clustering algorithm (DBSCAN) is applied for denoising, further reducing noise introduced by other highly reflective objects. Finally, precise alignment of reference points between two different coordinate systems is achieved through a robust iterative closest point (ICP) algorithm based on M-estimation.

The aforementioned method solves the problem of few common features between long-distance sensors by manually placing the calibration object within the common viewing area of multiple sensors \cite{wu2019points}. An alternative approach that does not use a calibration object but resolves the common viewing area issue through manual selection of features employs an improved scheme based on the Partial Iterative Closest Point (PA-PICP) algorithm. This method optimizes the registration process within the XY plane through time synchronization and keypoint selection. Another innovative aspect is the adjustment of the Z-axis using ground intersections. By identifying intersections on the ground as scanned by two different LiDARs, the method adjusts the Z-axis data to ensure vertical height consistency between the data collected by different sensors. This approach is particularly suitable for urban road environments with relatively flat terrain and can significantly improve the data alignment accuracy of roadside LiDAR systems.

\cite{lv2019revolution} utilizes ground point features and an innovative external parameter optimization search method based on rotation and revolution to achieve a fully automatic calibration process.

Beyond using traditional geometric features like points, lines, and planes, \cite{ren2023trajmatch} leverages trajectory information provided by detection and tracking modules for moving objects to identify corresponding trajectories between different LiDARs, thereby achieving calibration among them. During the calibration process, TrajMatch also considers semantic information, using semantically aware location matching to filter incorrect matches and enhance calibration accuracy. Moreover, TrajMatch introduces a continuous calibration method, which optimizes calibration parameters over multiple sessions, ultimately achieving centimeter-level spatial calibration accuracy and millisecond-level temporal calibration accuracy.

\cite{zhao2023spatial} proposes a spatial alignment framework utilizing geolocation cues, primarily applied in roadside multi-view multi-sensor fusion systems. Initially, a deep learning-based monocular localization method is used to estimate angles and distances in single-camera scenarios, inferring the camera's geographic location. Subsequently, a pseudo-camera parameter approximation approach is employed, using a minimal amount of LiDAR data to optimize the positioning results between multiple cameras, ensuring high precision in spatial alignment.

\subsection{Multi-end Dynamic Calibration}
\label{multiendDynamicCalibration}
The concept of multi-node dynamic calibration is aimed at sensors whose positions vary, such as vehicular sensors coordinating between infrastructure or other vehicles. As the relative positions of these sensors constantly change within such scenarios, the calibration process must adapt dynamically to ensure real-time data accuracy and reliability. The primary challenges faced by dynamic calibration include significant real-time changes in relative distances, notable variations in environmental features, and strong coupling with upstream and downstream systems. Currently, dynamic calibration technologies are primarily used for multi-node tasks in lidar sensors, where most calibration tasks rely on initial values provided by GPS/IMU. There is a trend of shifting from traditional independent methods towards more integrated and modular approaches. This includes the integration of calibration results into the perception system to enhance the accuracy of perception and the overall performance of the system. Within this broad category, methods are further distinguished between explicit and implicit relative position calibration techniques.



\begin{table*}[t]
\centering
\caption{SUMMARY OF EXPLICIT EXTRINSICS METHODS.}
\begin{tabular}{|c|c|c|c|c|}
\hline
\textbf{Method} & \textbf{Year} & \textbf{Sensor type \& Scenario} & \textbf{Feature type} & \textbf{Matching \& Optimization} \\
\hline
ISPRS\cite{yuan2022leveraging} & 2022 & LiDAR (Veh) + LiDAR (Veh) & point feature & RANSAC \\
\hline
IV\cite{song2023cooperative} & 2023 & LiDAR (Veh) + LiDAR (Veh) & geometry feature of Bounding Box & Optimal Transport Theory + RANSAC \\
\hline
RAL\cite{song2024spatial} & 2024 &LiDAR (Veh) + LiDAR (Veh) & geometry feature of Bounding Box & Local Context Matching + Global Consensus Maximization\\
\hline
ICRAS\cite{lu2023robust} & 2023 & LiDAR (Veh) + LiDAR (Veh) &geometry feature of Bounding Box &Graph Optimization \\
\hline
IV\cite{allig2019alignment} & 2019 & Veh + Veh & motion feature & Motion Fusion\\
\hline
CC\cite{duan2021v2i} & 2021 & LiDAR (Veh) - LiDAR (Inf) & point cloud map & NDT \cite{biber2003normal} \\
\hline
ACM\cite{he2021vi} & 2021 & LiDAR (Veh) - LiDAR (Inf)  & point-feature + line-feature & Ground Registration + Extended RANSAC Registration \\
\hline
IV\cite{zhao2023hpcr} & 2023 & LiDAR (Veh) - LiDAR (Inf) & points in roi region & FGR \cite{zhou2016fast} \\
\hline
ACM\cite{shi2022vips} & 2022 & LiDAR (Veh) - LiDAR (Inf) & geometry feature of Bounding Box & Graph Optimization \\
\hline
VTC\cite{maruta2021blind} & 2021 & LiDAR (Veh) - LiDAR (Inf) & motion feature & / \\
\hline
\end{tabular}
\label{tab:multi-end_dynamic_explicit_calibration}
\end{table*}

\subsubsection{Explicit Extrinsics Calibration}

Explicit methods treat calibration as an independent processing step, directly utilizing sensor data to output external parameters between devices. These parameters are subsequently used for downstream perception tasks, suitable for applications requiring high calibration accuracy and repeatability. These calibration methods are characterized by their independence and clarity, facilitating system debugging and parameter optimization.

In the process of solving for external parameters, the most direct method is to compute using motion data \cite{maruta2021blind}. This approach intuitively resolves the issue of the sensor data's common viewing area and possesses high versatility. In \cite{allig2019alignment}, the method employs unanticipated states of the sender for coordinate transformation and disregards sender motion compensation. Simulation comparisons of the two methods revealed that using the sender's unanticipated state more effectively reduces uncertainty and improves the accuracy of position estimates when processing received perception data. However, this approach has limitations, including significant errors often associated with motion data collection and the necessity for the chosen motion model to adapt to specific application scenarios.

Another traditional method involves the feature extraction-feature matching framework for external parameter computation, where features specifically refer to geometric features such as points, lines, and planes. Feature extraction typically falls into two categories: self-designed feature extractors and perception methods tailored for traffic elements. Self-designed feature extractors, such as the keypoint extraction algorithm proposed in \cite{he2021vi} and the line feature extraction algorithm proposed in \cite{sheng2024rendering}, have the advantage of customization, enabling deep integration with algorithms and enhancing overall system performance, though their general applicability still requires further experimental verification. On the other hand, feature extractors for traffic element perception, such as the ground extraction and lane detection algorithms in \cite{he2021vi}, object detection algorithm in \cite{yuan2022leveraging}, and image line feature extraction algorithm based on SAM in \cite{sheng2024rendering}, benefit from their modular structure's decoupling, offering greater flexibility and maintainability. As for feature matching, it primarily relies on mathematical algorithms for analyzing geometric features like points, lines, and planes, particularly point feature matching, which essentially simplifies the point cloud matching problem. It is noteworthy that these methods sometimes combine multiple features to meet the needs of different components of external parameters. For example, ground features in \cite{he2021vi} are used to align the z-axis degree of freedom, while the remaining five degrees of freedom are constrained by keypoints and lane features. This practice enhances the method's robustness and adaptability to various scenes.

As research in vehicle-to-infrastructure cooperation is still in its early stages, much of the current work continues to focus on the calibration among multi-node LiDARs. These studies primarily explore how to integrate point cloud registration issues with traffic scenarios. The paper \cite{duan2021v2i} demonstrates a map-based registration method for multi-LiDAR sensor devices in real-time applications. This method uses digital maps to perform real-time, high-precision registration of point clouds generated by multi-node LiDARs, ensuring the quality and availability of sensor data. However, this approach faces several challenges, including the collection and maintenance of the point cloud maps themselves and the selection issues between local point clouds and global maps.

Another mainstream method involves using perception results as constraints to select partial point clouds within the Region of Interest (ROI) for matching, with the key challenge being the precise acquisition of matched perception results. The studies \cite{song2023cooperative} and \cite{song2022efficient} utilize optimal transport theory to construct a cost matrix and derive an assignment matrix to determine the best matching pairs. To reduce errors in the perception layer and decrease computation, this method samples matching pairs and ultimately employs a method based on singular value decomposition for external parameter search. Additionally, the study \cite{song2024spatial} builds a context matrix based on the relative positions and other features of surrounding objects, providing a unique identifier for each object in its local scene. Based on this context matrix, the method calculates similarities between objects detected by different agents to obtain preliminary matching pairs, which are then refined through initial filtering and maximization of global consensus to determine the final matching pairs, using \cite{zhang2021fast} to solve for external parameters in the ROI area. Similarly, the paper \cite{shi2022vips} uses the location, semantics, and dimensions of detection frames to construct an affinity matrix, encoding a simplified representation of detection frames and their interrelations, and employs an efficient graph-structure matching algorithm \cite{leordeanu2005spectral} to acquire matching pairs and perform external parameter computations. Although the aforementioned methods utilize relatively coarse-grained perception technology like object detection, they introduce various additional matching strategies to further differentiate perception detection frames to obtain matching pairs, achieving commendable object matching efficiency. In contrast, the study \cite{zhao2023hpcr} employs fine-grained perception methods such as semantic segmentation and instance segmentation for direct matching. While this approach faces challenges with real-time performance, it offers another potential solution.

\begin{table*}[t]
\centering
\caption{SUMMARY OF IMPLICIT RELATIVE POSITIONING METHODS.}
\begin{tabular}{|c|c|c|c|c|}
\hline
\textbf{Method} & \textbf{Year} & \textbf{Sensor type \& Scenario} & \textbf{Feature type} & \textbf{Spatial alignment module} \\
\hline
ICDCS\cite{chen2019cooper} & 2019 & LiDAR (Veh) - LiDAR (Veh) & points in roi region & Spatial Feature Fusion \\
\hline
SEC\cite{chen2019f} & 2019 & LiDAR (Veh) - LiDAR (Veh) & Voxel feature & Spatial Feature Fusion/Voxel Feature Fusion \\
\hline
ECCV\cite{xu2022v2x} & 2022 & LiDAR (Veh) - LiDAR (Veh/Inf) & cnn feature & Spatial-Temporal Correction Module \\
\hline
ECCV\cite{wang2020v2vnet} & 2022 & LiDAR (Veh) - LiDAR (Veh) & cnn feature & Spatially-aware GNN \\
\hline
CoRL\cite{vadivelu2021learning} & 2020 & LiDAR (Veh) - LiDAR (Veh) & cnn feature & CNN perform pose regression + Attention Denoising Module \\
\hline
RAL\cite{yuan2022keypoints} & 2022 & LiDAR (Veh) - LiDAR (Veh) & point feature + Bounding Box & Robust network structures fpv-rcnn \\
\hline
TIV\cite{lin2024v2vformer} & 2024 & LiDAR (Veh) - LiDAR (Veh) & voxel-feature & Spatial-Aware Transformer \\
\hline
ICCV\cite{yang2023spatio} & 2023 & LiDAR (Veh) - LiDAR (Veh/Inf) & cnn feature & Spatio-temporal Feature Integration \\
\hline
IROS\cite{glaser2021overcoming} & 2021 & Camera (Drone) - Camera (Drone) & semantic feature & Multi-Agent Spatial Handshaking module \\
\hline
ICPADS\cite{shi2023mcot} & 2023 & Camera (Veh) - LiDAR (Veh) & cnn feature & BEV Feature Alignment with average sampling mechanism \\
\hline
ICCV\cite{xiang2023hm} & 2023 & Camera (Veh) - LiDAR (Veh) & cnn feature & Graph-structured feature fusion \\
\hline
\end{tabular}
\label{tab:multi-end-dynamic--calibration}
\end{table*}

\subsubsection{Implicit Extrinsics Calibration}

Implicit methods integrate the calibration process within perception tasks, treating calibration not as an independent output but as an intermediary variable dynamically adjusted within the perception algorithms. This approach significantly enhances system adaptability to dynamic environmental changes and is particularly suitable for dynamic scenarios that require real-time sensor configuration adjustments. Fundamentally, perception methods involve the organization and selection of spatial data, closely related to the spatial data alignment operations in calibration tasks. Traditional processes have not deeply explored the potential connections between calibration and perception tasks since sensor calibration can achieve high accuracy through manual operations. However, as the complexity of multi-node calibration tasks increases, more perception methods in practice have incorporated spatial alignment operations into their calibration. This paper attempts to revisit these multi-node perception methods' spatial alignment modules from the perspective of calibration tasks.

Traditional perception methods usually operate on the premise that precise external parameter values have already been obtained, and these methods are highly sensitive to deviations in these parameters. In static scenarios where sensor relative positions remain fixed, high-precision manual calibration methods can meet the needs of these traditional perception approaches. However, in multi-node dynamic scenarios, although traditional methods can obtain real-time pose information through GPS and IMU, estimating a rough set of external parameters, these parameters often have significant errors, significantly diminishing the effectiveness of traditional perception methods in these scenarios. Therefore, an important research direction is to incorporate spatial alignment operations into the spatial processing modules of perception methods to enhance robustness against disturbances in external parameters.

\cite{chen2019cooper} and \cite{chen2019f} propose two feature-based data fusion strategies, Voxel Feature Fusion (VFF) and Spatial Feature Fusion (SFF), to enhance the accuracy of object detection using 3D point cloud data under pose errors in autonomous driving vehicles. Voxel Feature Fusion (VFF) focuses on directly fusing voxel-level features between two vehicles, merging voxel features at the same location through a maxout function to improve data processing efficiency. Spatial Feature Fusion (SFF) involves preprocessing voxel features in each respective vehicle, extracting spatial features, and then performing fusion, allowing dynamic adjustment of the size of the transmitted feature maps to optimize communication efficiency. Both methods maintained good perception accuracy under external parameter errors of 0.2m on the T\&J dataset.

Additionally, some methods incorporate spatial alignment operations into classic perception networks. For instance, the FPV-RCNN proposed in \cite{yuan2022keypoints}, an extension based on PV-RCNN, primarily follows the design of PV-RCNN, including keypoint extraction and voxel feature fusion modules. The main innovation lies in introducing a generation and sharing module for the Current Position Module (CPM) to transfer real-time pose information, achieving high perception accuracy under localization noise on the COMAP, demonstrating high system robustness.

For instance, \cite{xu2022v2x} introduces V2X-ViT, which is based on the Vision Transformer (ViT) and customizes spatial-aware self-attention mechanisms and position encoding strategies to adapt to multi-node dynamic scenarios. Specifically, V2X-ViT incorporates a dynamic position encoding mechanism that can adjust the encoding dynamically based on the real-time motion state of the vehicle and surrounding environmental factors, allowing the model to update and adapt to changes in perspective and other environmental variations in real time. Moreover, V2X-ViT utilizes a spatial-aware self-attention mechanism, enhancing the model's perception capabilities through multi-scale feature fusion, enabling more effective handling of complex and dynamic traffic scenarios.

\cite{yang2023spatio} proposes SCOPE, a multi-agent collaborative perception framework. Its core is the Confidence-aware Cross-agent Collaboration (CCC) component based on the cross-attention mechanism, which effectively addresses the issue of misaligned feature maps among heterogeneous agents caused by localization errors. Similarly, \cite{glaser2021overcoming} introduces a multi-agent spatial interaction network (MASH), centered on a "spatial handshaking" process based on the cross-attention mechanism, allowing agents to identify and exchange image blocks with complementary visual information while maintaining low bandwidth consumption. To optimize feature matching and information fusion accuracy under localization errors, MASH further incorporates an autoencoder network to smooth and fill gaps in noisy matches and occluded areas.

In the processing of multimodal data, \cite{xiang2023hm} designs a spatial-aware heterogeneous 3D graph Transformer that effectively integrates features from LiDAR and cameras. HM-ViT introduces a graph structure to better represent the geometric relationships between objects, employing a Heterogeneous 3D Graph Attention Mechanism (H3GAT). Through this mechanism, the model can effectively capture spatial interactions and interrelations across agents while processing features generated by different types of sensors, which is particularly important in dynamic V2V perception due to frequent changes in relative positions and poses of vehicles.

In contrast to the ViT-based end-to-end network designs mentioned above, \cite{lin2024v2vformer} introduces a Transformer-based collaborative module (CoTr), which combines a Spatial-Aware Transformer (SAT) and a Channel-wise Transformer (CWT). SAT adjusts the importance of features based on the relevance of geographic locations, while CWT handles channel-level semantic interactions between vehicles and processes fused features through a feedforward network, achieving effective feature integration.

The geometric relationships of multiple targets in traffic scenarios are naturally suited to graph structures. \cite{xiang2023hm} integrates graph structures and attention mechanisms, while \cite{wang2020v2vnet} employs Graph Neural Networks (GNNs) to integrate information from multiple vehicles. It considers the relative positions and time delays between sending and receiving vehicles, using a spatial-aware messaging strategy to ensure spatially correct alignment of information. \cite{vadivelu2021learning} builds on V2VNet, optimizing predicted pose estimates through Markov Random Fields (MRF) and Bayesian reweighting and further using predicted attention weights to attenuate messages with high noise, ultimately achieving strong robustness against pose errors. As the aforementioned methods require extensive accurate pose ground truth data for training, which is costly to obtain in reality, \cite{lu2023robust} maintains high detection performance under ground truth poses with errors through innovative agent-object pose graph optimization techniques.

\subsection{Discussion}

The two classifications of multi-node dynamic calibration represent two different processing strategies. Explicit methods view calibration as an independent processing step, directly using sensor data to output external parameters between devices. These parameters are subsequently used for downstream perception tasks and are suitable for applications that require high calibration accuracy and repeatability. In contrast, implicit methods integrate the calibration process within perception tasks, where calibration is no longer an independent output but is dynamically adjusted as an intermediary within the perception algorithms. This approach enhances the system's adaptability to dynamic environmental changes and is particularly suitable for dynamic scenarios that require real-time adjustments to sensor configurations. The choice between these methods depends on specific system requirements and operational environments, with explicit methods emphasizing accuracy and independence, while implicit methods emphasize integration and dynamic adjustment capabilities.

Through these advanced calibration techniques, the collaborative efficiency and accuracy of multi-node sensor systems can be greatly enhanced, thereby advancing the application and development of intelligent systems in complex environments.

\begin{table*}[htbp]
\centering
\caption{MULTI-END CALIBRATION DATASET OVERVIEW.}
\begin{tabular}{|c|c|c|c|c|c|c|c|}
\hline
\multirow{2}{*}{\textbf{Paper}} & \multirow{2}{*}{\textbf{Year}} & \multirow{2}{*}{\textbf{Venue}} & \multirow{2}{*}{\textbf{Scenario}} & \multirow{2}{*}{\textbf{Source}} & \multicolumn{2}{c|}{\textbf{Sensor}} & \multirow{2}{*}{\textbf{Agents number}} \\
\cline{6-7}
& & & & & \textbf{LiDAR} & \textbf{Camera} & \\
\hline
V2V-Sim\cite{wang2020v2vnet} & 2020 & ECCV & V2V & Simulator (LiDARsim) & \cmark & \xmark  & 7 \\
\hline
CoopInf\cite{arnold2020cooperative} & 2020 & TITS & V2I & Simulator (CARLA) & \xmark & \cmark & 6,8 \\
\hline
CODD\cite{arnold2021fast} & 2021 & RAL & V2V & Simulator (CARLA) & \cmark & \xmark & 10 \\
\hline
V2X-Sim\cite{li2022v2x} & 2022 & RAL & V2V, V2I & Simulator (CARLA) & \cmark & \cmark & 5 \\
\hline
AUTOCATSIM\cite{cui2022coopernaut} & 2022 & CVPR & V2V & Simulator (CARLA) & \cmark & \cmark & - \\
\hline
CoopInfo\cite{arnold2020cooperative} & 2022 & TITS & V2I & Simulator (CARLA) & \cmark & \cmark & 6,7 \\
\hline
DOLPHINS\cite{mao2022dolphins} & 2022 & ACCV & V2V, V2I & Simulator (CARLA) & \cmark & \cmark & 3 \\
\hline
COMAP\cite{yuan2021comap} & 2022 & RAL & V2V & Simulator (CARLA, SUMO) & \cmark & \xmark & 10 \\
\hline
OPV2V\cite{xu2022opv2v} & 2022 & ICRA & V2V & Simulator (CARLA, OpenCDA) & \cmark & \cmark & 7 \\
\hline
V2X-Set\cite{xu2022v2x} & 2022 & ECCV & V2V, V2I & Simulator (CARLA, OpenCDA) & \cmark & \cmark & 5 \\
\hline
DeepAccident\cite{wang2024deepaccident} & 2023 & arXiv & V2V, V2I & Simulator (CARLA) & \cmark & \cmark & 5 \\
\hline
A9-Dataset\cite{cress2022a9} & 2022 & IV & V2I & Real-World (Munich, Germany) & \cmark & \cmark & 1 \\
\hline
LUCOOP\cite{axmann2023lucoop} & 2023 & IV & V2V & Real-World (Hannover, Germany) & \cmark & \xmark & 3 \\
\hline
DAIR-V2X\cite{yu2022dair} & 2022 & CVPR & V2I & Real-World (Beijing, China) & \cmark & \cmark & 2 \\
\hline
IPS300+\cite{wang2022ips300+} & 2022 & ICRA & V2I & Real-World (Beijing, China) & \cmark & \cmark & 2 \\
\hline
V2X-Seq\cite{yu2023v2x} & 2023 & CVPR & V2V, V2I & Real-World (Beijing, China) & \cmark & \cmark  & 2 \\
\hline
T\&J\cite{chen2019cooper} & 2019 & ICDCS & V2V & Real-World & \cmark & \xmark  & 2 \\
\hline
WIBAM\cite{howe2021weakly} & 2021 & BMVC & V2I & Real-World & \xmark & \cmark  & 4 \\
\hline
V2V4Real\cite{xu2023v2v4real} & 2023 & CVPR & V2V & Real-World & \cmark & \cmark  & 2 \\
\hline
\end{tabular}
\label{tab:multi-end-static-calibration}
\end{table*}

\section{Dataset \& Application and Future Perspectives}
\subsection{Dataset}
Datasets play a crucial role in multi-sensor calibration tasks. These datasets typically require data from corresponding sensors and provide precise relative pose information. Given the specificity of calibration tasks, the necessary datasets can generally be sourced from various perception tasks. These datasets can be categorized into real-world collected datasets and simulated datasets generated by simulators.

In terms of simulated datasets, the CARLA simulator \cite{dosovitskiy2017carla} is commonly used to generate environments that can provide very accurate extrinsic parameters, which are beneficial for researching and developing complex perception algorithms. CARLA is particularly suitable for autonomous driving research as it offers detailed simulations of roads, traffic, and weather conditions. Other commonly used simulators include LidarSim \cite{manivasagam2020lidarsim} for LiDAR sensors, SUMO \cite{krajzewicz2010traffic} for urban traffic scenarios, and the OpenCDA \cite{xu2021opencda} framework for multi-vehicle collaboration. These simulation software can sometimes be used in combination to achieve better specific effects.

For real-world datasets, extrinsic parameters are typically obtained through measurement and annotation, and these parameters inevitably contain certain errors. The literature \cite{liu2023towards} reflects the quality differences in dataset calibration based on experiments with common multi-sensor perception methods on multi-sensor datasets. As discussed in section\ref{multiendDynamicCalibration}, methods for multi-sensor calibration tend to integrate spatial alignment with perception techniques. Practically, this manifests as evaluating perception methods under noisy pose conditions. On this level, the real-world complexity and noise in the extrinsic parameters of real datasets actually become an advantage, allowing for a better evaluation of the effectiveness of these calibration modules.

From the perspective of sensor types, although existing multi-sensor calibration methods primarily target multi-LiDAR systems, many current multi-sensor datasets also include data from heterogeneous sensors such as cameras. This indicates that calibration tasks for heterogeneous sensors are well-supported at the data support level.

Regarding the datasets' scenarios, many designs include both vehicle-to-vehicle (V2V) and vehicle-to-infrastructure (V2I) scenarios. The multi-sensor calibration methods mentioned in section \ref{multiendCalibration} generally have certain applicability in both scenarios \cite{yang2023spatio}\cite{xu2022v2x}. Comparatively, the co-visibility areas between different endpoints in V2V scenarios change more drastically; whereas in V2I scenarios, the fixed position of roadside sensors provides a more stable global view, ensuring a certain extent of data co-visibility. Furthermore, since most multi-sensor calibration methods rely on initial pose values provided by positioning systems, this significantly reduces the difficulty of locating co-visibility areas, making calibration tasks in these scenarios simpler and minimizing differences between the two scenarios.

\subsection{Application}
Collaboration between Intelligent Vehicles and Road Infrastructure (V2I) refers to the communication and exchange of perception or positional information between intelligent vehicles and road infrastructure. Despite the advancements in autonomous driving, the resolution of long-distance objects within the perceptual information of individual vehicles remains comparatively low and is often overlooked. Additionally, due to limited viewpoints, individual perception images suffer from significant occlusion issues. Challenges such as lower accuracy in vehicle localization, stringent real-time requirements, and the need to address occlusions have led to the rapid development of cooperative perception in intelligent connected vehicles. In the context of autonomous driving, cooperative interactions primarily occur among intelligent vehicles (V2V), between intelligent vehicles and road infrastructure (V2I), and among intelligent vehicles and other entities (V2X). With known relative pose relationships between entities, intelligent vehicles can effectively acquire information from neighboring vehicle sensors or roadside devices. This information is crucial for tasks like path planning, collision avoidance, and vehicle control. A cooperative target detection method for vehicle-infrastructure interactions was proposed by Yu et al.\cite{yu2022multistage}, while Zhang et al. \cite{zhang2022multi} utilized a heterogeneous graph attention network to enhance the fusion of interactions among intelligent agents. Addressing challenges in vehicle-infrastructure cooperative tasks, Yu et al.\cite{yu2023vehicle} employed a feature flow prediction module to tackle time asynchrony in traffic environments and potential fusion misalignment caused by limited wireless communication. Furthermore, Song et al.\cite{song2023cooperative} unified multi-end perception information into a coordinated system and introduced two matching algorithms based on Iterative Closest Point (ICP) and optimal transport theory, significantly amplifying vehicle localization accuracy. The diverse perspectives from adjacent vehicles or roadside devices compensate for perceptual blind spots of the current vehicle. For instance, overhead views from intersection devices can offer information about occluded vehicles or pedestrians, while perspectives from other vehicles around corners can provide insight into blind spots for the current vehicle at intersections.

The accurate relative pose between intelligent vehicles and road infrastructure is a prerequisite for cooperative between the vehicle and the infrastructure. Cooperative interaction can address certain challenges that individuals cannot handle, such as occlusion, long-distance perception, and vehicle positioning offset. In V2I cooperative scenarios, the positioning information provided by the infrastructure typically exhibits higher accuracy and stability. Single vehicles can leverage the auxiliary information from the infrastructure to optimize their positioning algorithms, thereby enhancing positioning accuracy and reliability. Infrastructure sensors enable cooperative perception among vehicles, allowing different vehicles to share environmental information. Through vehicle cooperative perception, single vehicles can better cope with complex traffic conditions, thereby improving traffic efficiency and safety.

\subsection{Future Perspectives}
The precise calibration of sensor relative poses between infrastructure and autonomous vehicles holds immense potential in the realm of advanced transportation research. By establishing accurate inter-sensor alignment, this calibration plays a pivotal role in bolstering various aspects of vehicle-assisted positioning and cooperative perception. Through this calibration, not only can auxiliary positioning information be provided to the autonomous vehicles, but it also has the capacity to amplify their environmental perception capabilities. The calibrated cooperation between sensors on both the infrastructure and the vehicles can lead to a substantial improvement in the vehicles' awareness of their surroundings, thus increasing overall road safety.

The notion of vehicle-road collaboration, where vehicles seamlessly share perception data with infrastructure and each other, becomes more attainable through well-calibrated sensor setups. This cooperative perception can lead to a more comprehensive understanding of the driving environment, contributing to the realization of efficient and informed traffic flow management. This is particularly vital in congested urban environments, where congestion mitigation and efficient navigation are paramount.

As we look forward, it is apparent that the ongoing synergy between advanced technologies and intelligent transportation systems will continue to reshape the landscape of single vehicle travel. The ongoing advancements in sensor calibration techniques, coupled with the ever-increasing capability of intelligent infrastructure, promises to usher in a new era of smarter and safer transportation experiences. This pursuit of innovation not only furthers the domain of autonomous vehicles but also contributes significantly to the broader goal of creating sustainable and intelligent transportation networks. Through collaborative research and technological evolution, we can anticipate a future where single vehicle travel is not only more convenient but also contributes to a safer and more sustainable urban ecosystem.

\subsection{Conclusion}
Camera-LiDAR calibration has always been a key aspect in the field of autonomous driving technology and has received a great deal of attention from the industry. With the emergence of the concept of multi-terminal cooperation, where infrastructure assists the vehicle by transferring environmental data, enhancing its sensing capabilities, and reducing costs, a new and complex technology area has emerged. The importance of inter-sensor calibration cannot be overemphasized especially in scenarios where multiple sensors are collaboratively involved in sensing tasks and localization challenges. Diverse end-point calibration strategies urgently need to be improved.

In the field of autonomous driving, cameras and LiDAR sensors are cornerstone elements. Their wide range of applications and critical role in perception make them the focus of calibration efforts. This paper presents a comprehensive analysis of multi-terminal camera-LiDAR calibration, delving into viewpoints that include vehicle-side, road-side, and vehicle-road cooperative perspectives. These calibration paradigms are explored in depth by outlining their respective applications and deeper implications.

Finally, a brief summary is provided, extrapolating into the future and looking at the trajectory of sensor calibration in the field of autonomous driving. As technology advances and the integration of vehicles and infrastructure becomes more common, calibration methods will continue to evolve to meet these needs. This exploration emphasizes the importance of the continued evolution of calibration as a key enabler for achieving safe and efficient autonomous driving systems.

\vfill

\end{document}